\documentclass{article}

\usepackage[preprint]{neurips_2024}

\usepackage[utf8]{inputenc} 
\usepackage[T1]{fontenc}    
\usepackage{hyperref}       
\usepackage{url}            
\usepackage{booktabs}       
\usepackage{amsfonts}       
\usepackage{nicefrac}       
\usepackage{microtype}      
\usepackage{xcolor}         
\usepackage{amsmath}
\usepackage{subcaption}

\usepackage{graphicx}
\usepackage{bm}
\usepackage{amssymb}
\usepackage{algorithm}
\usepackage{algpseudocode}
\usepackage{multirow}
\title{Towards Stable and Storage-efficient Dataset Distillation: Matching Convexified Trajectory}

\author{
Wenliang Zhong  \\
School of software\\
Shandong University\\
\And
Haoyu Tang \thanks{ Tang Haoyu is the corresponding author; email: zhongwenliang@mail.sdu.edu.cn, tanghao258@sdu.edu.cn} \\
School of software\\
Shandong University\\
\AND
Qinghai Zheng \\
College of Software \\
Fuzhou University \\
\\
\And
Mingzhu Xu \\
School of software\\
Shandong University\\
 \\
\And
Yupeng Hu \\
School of software\\
Shandong University\\
\\
\And
Liqiang Nie \\
School of Computer Science\\
Harbin Institute of Technology (Shenzhen) \\
 \\
}

\begin{document}

\maketitle

\begin{abstract}
The rapid evolution of deep learning and large language models has led to an exponential growth in the demand for training data, prompting the development of Dataset Distillation methods to address the challenges of managing large datasets. Among these, Matching Training Trajectories (MTT) has been a prominent approach, which replicates the training trajectory of an expert network on real data with a synthetic dataset. However, our investigation found that this method suffers from three significant limitations: 1. Instability of expert trajectory generated by Stochastic Gradient Descent (SGD); 2. Low convergence speed of the distillation process; 3. High storage consumption of the expert trajectory. To address these issues, we offer a new perspective on understanding the essence of Dataset Distillation and MTT through a simple transformation of the objective function, and introduce a novel method called Matching Convexified Trajectory (MCT), which aims to provide better guidance for the student trajectory. MCT leverages insights from the linearized dynamics of Neural Tangent Kernel methods to create a convex combination of expert trajectories, guiding the student network to converge rapidly and stably. This trajectory is not only easier to store, but also enables a continuous sampling strategy during distillation, ensuring thorough learning and fitting of the entire expert trajectory. Comprehensive experiments across three public datasets validate the superiority of MCT over traditional MTT methods. 
\end{abstract}

\section{Introduction}
The advancement of deep learning has catalyzed an exponential surge in the requisite volume of training data \citep{wang2022cafe}. With the emergence of Large Language Models (LLMs), there has been a corresponding rise in model complexity, further intensifying the demand for extensive datasets to facilitate the training of these intricate models. However, collecting and managing large datasets presents significant challenges, including storage requirements, computational load, privacy concerns, and the costs of data labeling. To mitigate these challenges, Dataset Distillation (DD) has emerged as a compelling strategy \citep{wang2018dataset}. DD endeavors to distill the essence of a large, real-world dataset into a more compact, synthetic dataset that can train models with comparable efficacy.

In the landscape of DD methods, Matching Training Trajectories has emerged as a prominent approach. The MTT method aims to generate a synthetic dataset that guidea the learning trajectory of the student network to approximate the expert trajectory of this network on real data. However, upon closer examination, we identify several limitations inherent in traditional MTT approaches:

\begin{figure}
  \centering
  \begin{subfigure}{0.36\linewidth}
    \includegraphics[width=\linewidth]{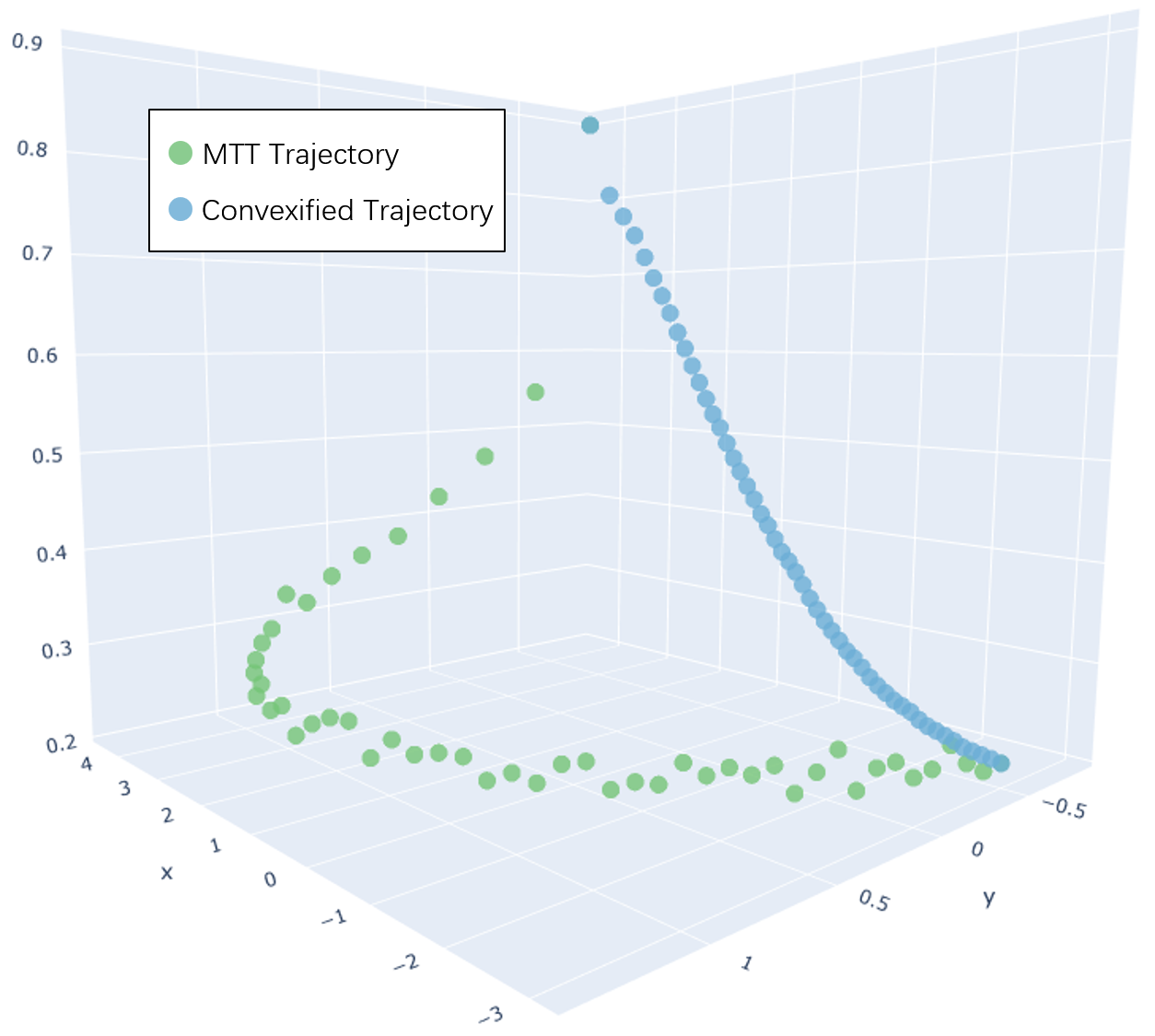}
    \caption{Visualization of the expert trajectory.}
    \label{fig:intro_visual}
  \end{subfigure}
  \begin{subfigure}{0.36\linewidth}
    \includegraphics[width=\linewidth]{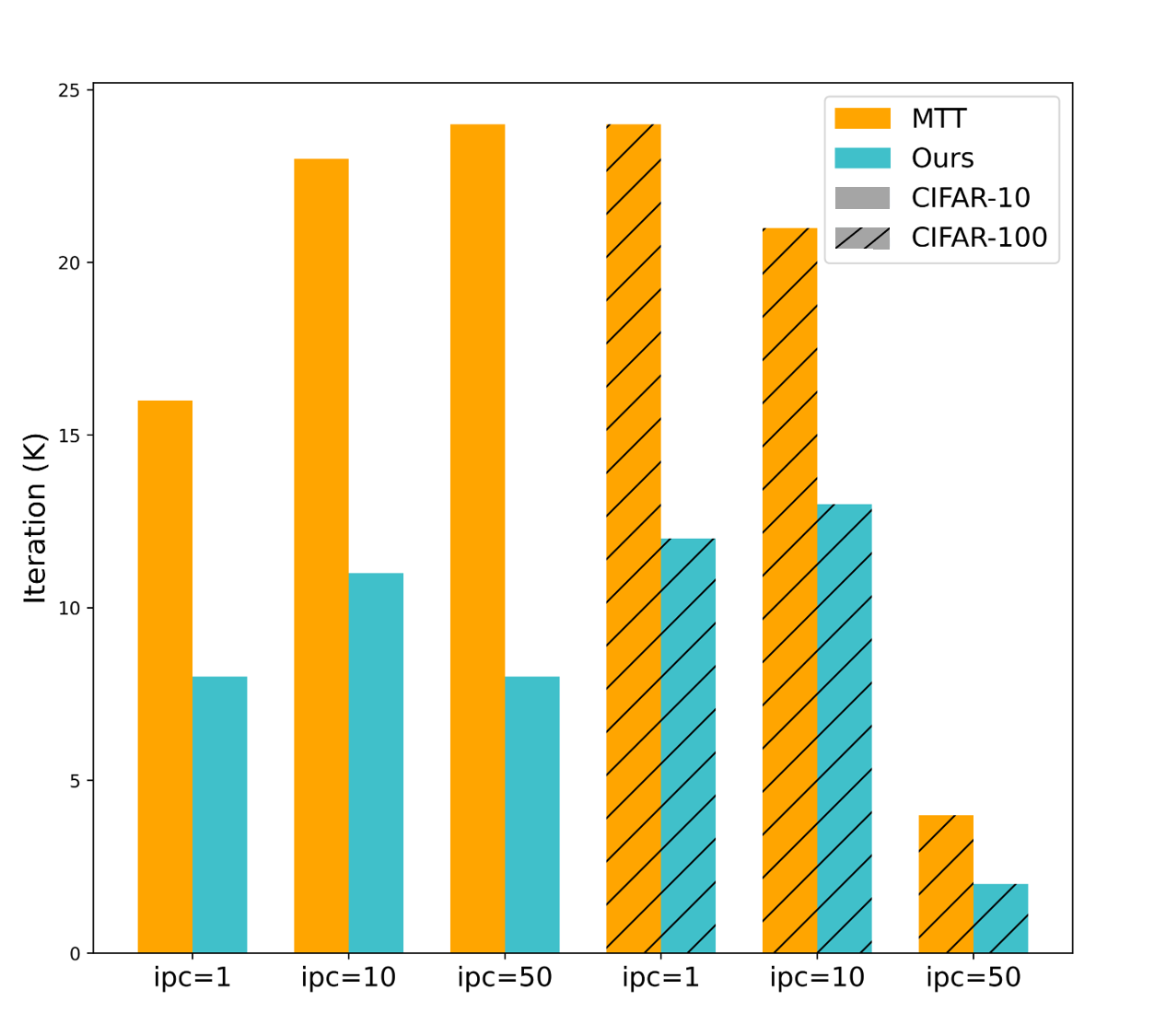}
    \caption{Illustration of Convergence Speed}
    \label{fig:intro_speed}
  \end{subfigure}

  \caption{(a): PCA projection of all waypoints model in the expert trajectory, where z-axis represents the value of $(1-\text{validation accuracy})$; (b): The required iteration number to convergence for both the MCT and MTT methods during distillation. The convergence is defined by the condition where the difference between the accuracy at any iteration and the maximum accuracy is less than $\epsilon$=2\%.}
  \label{fig:intro}
\end{figure}
1. \textbf{Instability of expert trajectory}: As shown in Figure \ref{fig:intro_visual}, the validation accuracy of the expert network on the MTT trajectory exhibits oscillations. Matching the trajectory locally in each iteration will lead to the similar oscillation in the trajectory of the synthetic data, thereby impeding robust distillation.

2. \textbf{Low Convergence Speed:} The learning process for the expert trajectory is often slow. As in Figure \ref{fig:intro_speed}, a considerable number of distillation iterations are required to generate a synthetic dataset capable of achieving satisfactory test accuracy, resulting in time-consuming procedures.

3. \textbf{High Storage Consumption}: During the distillation process, the conventional MTT approach necessitates the storage of model weights along all timesteps, which is particularly burdensome in terms of storage (about 50 models should be stored). This high storage consumption is a significant limitation for applying existing DD methods to small-scale models.

Through careful observation, we have reformulated the loss function of MTT, and introduced a novel perspective to interpret the essence of DD and MTT: obtaining a synthetic dataset that offers accurate guidance regarding the magnitude and direction of the next update for any given point in the parameter space of the student model, with this guidance determined by the expert trajectory's update vector at that point. From this perspective, those three limitations can be easily addressed: to find an optimized expert trajectory that can guide the model to stably converge at each iteration, which is also easy to fit and simple to save.

How to find such a trajectory? Drawing inspiration from linearized dynamics of Neural Tangent Kernel (NTK) method \citep{arora2019exact,jacot2018neural}, we present a simple yet novel Matching Convexified Trajectory (MCT) method. The MCT method creates a convex combination (linear) expert trajectory based on the network's training process real data. This trajectory, which starts from a random initialization model and points directly towards the optimal model point, facilitates stable and rapid convergence of the distillation. Moreover, recovering this trajectory only needs storing two models and a set of constants. Distinct from the MTT method, the convexified trajectory also permits a “continuous sampling” strategy during the distillation, ensuring comprehensive learning and fitting of the expert trajectory. 

The contributions of this paper are as follows: 1) We highlight the three limitations of traditional MTT methods, and offer a novel perspective for understanding the objective of DD through a simple reformulation of MTT's loss function. 2) We propose the MCT method, which creates an easy-to-store convexified expert trajectory with a continuous sampling strategy to enable rapid and stable distillation. 3) Comprehensive experiments on three datasets have verified the superiority of our MCT and the effectiveness of the continuous sampling strategy.

\section{Preliminaries and Related Work}

\subsection{Preliminaries}
We first formally define the dataset distillation task. A large scale real dataset $\mathcal{T}=\{(x^{(i)}_{\mathcal{T}}, y^{(i)}_{\mathcal{T}})\}_{i=1}^{|\mathcal{T}|}$ is first provided, where $x^{(i)}_{\mathcal{T}} \in \mathbb{R}^{d}$ and $y^{(i)}_{\mathcal{T}} \in \mathcal{Y}=\{1,2,\dots,C\}$ are the $i$-th instance and the corresponding label. $C$ denotes the number of classes. The core idea of this task is to learn a tiny synthetic dataset $\mathcal{S}=\{(x^{(i)}_{\mathcal{S}}, y^{(i)}_{\mathcal{S}})\}_{i=1}^{|\mathcal{S}|}$ from the original dataset $\mathcal{T}$, where $x^{(i)}_{\mathcal{S}} \in \mathbb{R}^{d}$ and $y^{(i)}_{\mathcal{S}} \in \mathcal{Y}$. Typically, $ipc$ instances are crafted for each class, culminating in a total count for $\mathcal{S}$ of ${|\mathcal{S}|}=C*ipc$. It is always expected that ${|\mathcal{S}|} \ll {|\mathcal{T}|}$, while $\mathcal{S}$ still preserves the majority of the pivotal information in $\mathcal{T}$. Consequently, a model trained on  $\mathcal{S}$ should achieve performance comparable to the model trained with the original dataset $\mathcal{T}$ under the real data distribution $\mathcal{P}_{D}$. Formally, the optimization of DD task can be formulated as: 
\begin{equation}
    \arg\min_{\mathcal{S}}\mathcal{L}(\mathcal{S},\mathcal{T}),
\end{equation}
where $\mathcal{L}$ is the certain objective function, which may differ from different DD methods. 

\subsection{Dataset Distillation Methods.}
The field of DD contains four principal approaches. 
\textit{a. \textbf{Meta-model Matching methods}} \citep{wang2018dataset, zhou2022dataset, nguyen2021dataset, loo2022efficient} involve a bi-level optimization algorithm where the inner loop updates the weights of a differentiable model using gradient descent on a synthetic dataset while caching recursive computation graphs, and the outer loop validates models trained in the inner loops on a real dataset, back-propagating the validation loss through the unrolled computation graph to the synthetic dataset. \textit{b. \textbf{Distribution Matching methods }} \citep{zhao2023dataset,wang2022cafe} align synthetic and real data by optimizing within a set of embedding spaces using maximum mean discrepancy. However, inaccurate estimation of the data distribution often results in suboptimal performance. \textit{c. \textbf{Single-step Gradient Matching methods}} \citep{zhao2020dataset,zhao2021dataset} aim to align the gradient of the synthetic dataset with that of the real dataset during each training step. To enhance generalization with improved gradients, recent research efforts have focused on further optimizing the gradient matching objective by incorporating class-related information \citep{lee2022dataset,jiang2023delving}. \textit{d. \textbf{Multi-step Trajectory Matching methods}} \citep{cazenavette2022dataset,guo2023towards} address the accumulated trajectory errors of single-step methods by matching the multi-step training trajectories of models separately trained on synthetic and real datasets.

Our research primarily focuses on multi-step trajectory matching methods. The first method in this branch is MTT \citep{cazenavette2022dataset}. 
Based on MTT, \citet{du2023minimizing} presented to incorporate the random noise to the initialized model weights to mitigate accumulated trajectory errors, and \citet{cui2023scaling} proposed to decompose the objective function of MTT to improve computational efficiency and reduce GPU memory without performance degradation. 
Further research has explored the robustness of the synthesized dataset \citep{guo2023towards,li2022dataset,du2024sequential} and applied this technique to downstream tasks \citep{li2022compressed,li2020soft}.

Despite their successes, none of these approaches address the detriment of oscillations in the MTT expert trajectory on the stability and convergence speed of the distillation process. Furthermore, the necessity to retain all waypoint networks along the expert trajectory has yet to be addressed.

\section{Motivation}

\subsection{Review of Multi-step Trajectory Matching} 
In this section, we first review the multi-step trajectory matching methods. The essence of them is to minimize the discrepancy of the student training trajectory of $\mathcal{S}$ and the expert training trajectory of $\mathcal{T}$. Here we take MTT \citep{cazenavette2022dataset} as an example. Firstly, an expert trajectory $\tau_{\text{mtt}}=\{\theta_{\mathcal{T}}^{(t)}|0\leq t\leq K\}$ is generated by training a randomly initialized model $\theta_{\mathcal{T}}^{(0)}$ on the real dataset $\mathcal{T}$ with $K$ timesteps. Afterward, MTT matches the student trajectory with the expert $\tau_{\text{mtt}}$ through massive iterations. During each iteration, MTT samples a random timestep $\theta^{(t)}_\mathcal{T}$ and captures the target timestep $\theta^{(t+M)}_{\mathcal{T}}$ after $M$ steps from $\tau_{\text{mtt}}$. Meanwhile, $\theta^{(t)}_\mathcal{T}$ is also trained on synthetic dataset $\mathcal{S}$ for $N$ steps to get the updated student parameters $\theta^{(t+N)}_\mathcal{S}$. Formally, the objective is to minimize the normalized squared $L_2$ error between the updated student parameters $\theta^{(t+N)}_\mathcal{S}$ and the future expert (target) parameters $\theta^{(t+M)}_{\mathcal{T}}$: 
\begin{equation}
    \label{equ:MTT_loss}
    \mathcal{L}(\mathcal{S},\mathcal{T}) = \frac{\|\theta^{(t+N)}_{\mathcal{S}} - \theta^{(t+M)}_{\mathcal{T}}\|^2_2}{\|\theta^{(t)}_{\mathcal{T}} - \theta^{(t+M)}_{\mathcal{T}}\|^2_2} 
\end{equation}
\begin{equation}
\theta_{\mathcal{S}}^{(t+1)} = \theta_{\mathcal{S}}^{(t)} - \alpha_\mathcal{S} \nabla \ell (\mathcal{S}; \theta_{\mathcal{S}}^{(t)})
\end{equation}    
\begin{equation}
\theta_{\mathcal{T}}^{(t+1)} = \theta_{\mathcal{T}}^{(t)} - \alpha_\mathcal{T} \nabla \ell (\mathcal{T}; \theta_{\mathcal{T}}^{(t)}),
\end{equation}   
where $\theta_{\mathcal{T}}^{(t)}=\theta_{\mathcal{S}}^{(t)}$. $\ell$ is the loss function for model training, where the cross-entropy loss is often adopted, and $\alpha_\mathcal{S}$ and $\alpha_\mathcal{T}$ are the learning rates for training on the synthetic and real datasets, respectively. To ensure generalization, MTT usually performs the above trajectory matching process on a large number of expert trajectories from different $\theta_{\mathcal{T}}^{(0)}$. Although the subsequent methods have focused on optimizing model parameters \citep{du2023minimizing,du2024sequential} and objective functions \citep{cui2023scaling}, the overall process remains roughly the same as MTT.

\subsection{Motivation: A New Perspective to Optimize the Trajectory}
Through a lot of preliminary experiments and visualizations, we found that the MTT method have three serious shortcomings:
1. \textit{\textbf{Instability of the expert trajectory generated by mini-batch SGD:}} The expert trajectory $\tau_{\text{mtt}}$ trained on $\mathcal{T}$ exhibits erratic oscillations instead of following a path where the loss steadily decreases, so the accuracy of the waypoint model $\theta^{(t)}_\mathcal{T}$ is subject to fluctuations. This problem complicates the student network to learn better training dynamics with synthetic data. 2. \textit{\textbf{Low convergence speed of the distillation process:}} When learning expert trajectories, a very large number of iterations are required to obtain a synthetic dataset that can achieve good validation accuracy, which is very time-consuming. 3. \textit{\textbf{High storage consumption of the expert trajectory:}} To expedite the distillation process, the expert trajectories are pre-generated and stored in memory as trajectory buffers. These trajectories serve as sources from which the initial point $\theta^{(t)}_\mathcal{T}$ and target parameters $\theta^{(t+M)}_{\mathcal{T}}$ are extracted. However, the necessity to store all waypoints for each expert trajectory incurs a substantial storage footprint.



To better explain the internalization of these drawbacks, we propose a novel perspective to view the dataset distillation and explain the essence of the MTT approach: The objective of DD task can be regarded as obtaining a set of parameters (\textit{i.e.}, the synthetic dataset $\mathcal{S}$) that enables accurate prediction of how far (\textbf{magnitude}) and where (\textbf{direction}) to step next for any given network parameters $\theta$ (\textit{i.e.}, provides appropriate guidance $\Vec{V}_{\mathcal{S}}$ to update the current network parameters $\theta$). 
From this perspective, each distillation iteration of the MTT method can be viewed as updating the synthetic dataset $\mathcal{S}$ to provide the network update guidance $\Vec{V}_{\mathcal{S}}=\|\theta^{(t+N)}_{\mathcal{S}} - \theta^{(t)}_{\mathcal{T}}\|^2_2$ of $N$-step SGD training on $\mathcal{S}$, which aligns closer to the $M$-step SGD guidance $\Vec{V}_{\mathcal{T}}=\| \theta^{(t+M)}_{\mathcal{T}}-\theta^{(t)}_{\mathcal{T}}\|^2_2$ obtained from the expert trajectory, given an arbitrary initialized point $\theta_{\mathcal{T}}^{(t)}$. A simple reformulation of Equ.~\ref{equ:MTT_loss}  yields the same result:
\begin{equation}
        \min_{\mathcal{S}}\mathcal{L}(\mathcal{S},\mathcal{T}) 
        =\min_{\mathcal{S}} \mathbb{E}_{\theta^{(t)}_{\mathcal{T}} \sim \tau_\text{mtt}}  \frac{\|(\theta^{(t+N)}_{\mathcal{S}} - \theta^{(t)}_{\mathcal{T}})- (\theta^{(t+M)}_{\mathcal{T}}-\theta^{(t)}_{\mathcal{T}})\|^2_2}{\|\theta^{(t)}_{\mathcal{T}} - \theta^{(t+M)}_{\mathcal{T}}\|^2_2}
        =\min_{\mathcal{S}} \mathbb{E}_{\theta^{(t)}_{\mathcal{T}} \sim \tau_\text{mtt}} \frac{\left\|\Vec{V}_{\mathcal{S}}-\Vec{V}_{\mathcal{T}}\right\|^2_2}{\left\|\Vec{V}_{\mathcal{T}}\right\|^2_2}, \\   
\end{equation}
Thereafter, we can regard all the waypoints of $\tau_\text{mtt}$ as the training ``dataset'' to optimize $\Vec{V}_{\mathcal{S}}$, \textit{i.e.}, ${ \{ ( \theta^{(t)}_{\mathcal{T}}, \Vec{V}^{(t)}_{\mathcal{T}})| \theta^{(t)}_{\mathcal{T}} \in \tau_\text{mtt}, 0 \leq t \leq K \}}$, where $\Vec{V}^{(t)}_{\mathcal{T}}$  denotes the ``label'' of $\theta^{(t)}_{\mathcal{T}}$.
From this perspective, the first two drawbacks can be easily explained: Given that the models on the expert trajectory $\tau_\text{mtt}$ are all obtained by SGD training, and considering the variations in sample distribution across mini-batches, the expert trajectory $\tau_\text{mtt}$ has huge oscillations. Therefore, the training dynamics $\Vec{V}^{(t)}_{\mathcal{T}}$ obtained by sampling two arbitrary points with an interval of $M$ steps from $\tau_\text{mtt}$  cannot guarantee to always provide a favorable direction for $\Vec{V}^{(t)}_{\mathcal{S}}$ to learn. The final result is 1) poor $\Vec{V}^{(t)}_{\mathcal{T}}$ leads to instability; 2) considerable time is expended in identifying the optimal optimization direction to achieve convergence. This raises the question: Is there a superior trajectory $\hat{\tau}$ that consistently delivers more advantageous $\Vec{V}^{(t)}_{\mathcal{T}}$ to optimize the synthetic dataset $\mathcal{S}$ through $\Vec{V}^{(t)}_{\mathcal{S}}$?

We believe that an ideal expert trajectory should: 1) For any $\hat{\theta}_{\mathcal{T}}^{(t)}$ on $\hat{\tau}$, the obtained  $\Vec{V}^{(t)}_{\mathcal{T}}$ should always point to the direction that guides the target loss $\ell (\mathcal{T}; \theta_{\mathcal{T}}^{(t)})$ to decrease; 2) This trajectory is easier to fit for $\mathcal{S}$, because the size of $\mathcal{S}$ is much smaller than the original dataset $\mathcal{T}$. 3) The trajectory is easy to save and restore.

We draw inspiration from convex optimization \citep{boyd2004convex, bubeck2015convex} and NTK \citep{jacot2018neural, hanin2019finite}. First, since deep learning is essentially a non-convex problem, if we can make expert trajectories exhibit more convex properties, optimization becomes much less difficult. How to find convex trajectories? NTK methods prove that for a neural network $f_{\theta}(x)$, its update can be approximated by its first-order Taylor expansion in the neural network tangent space \citep{lee2019wide}: 
\begin{equation}
    f_{\theta}(x) \approx f_{lin,\theta}(x) = f_{\theta_0}(x) + (\theta-\theta_0)^{\mathsf{T}} \nabla_{\theta} f_{\theta_0}(x). 
\end{equation}

From this, we believe that replacing the original trajectory with a convex combination (linear) trajectory would be much more effective. The starting and ending points of this linear trajectory are the same as $\tau_\text{mtt}$, and all the waypoints are distributed along this line. This trajectory meets our needs very well: 1) The visualization in Figure \ref{fig:intro_visual} verifies that the validation accuracy of the model on this trajectory consistently increases; 2) The direction of any $\Vec{V}^{(t)}_{\mathcal{T}}$ sampled from this trajectory is always from the starting point to the ending (optimal) point, which is easy to fit for distilled data; 3) Only the parameters of its starting and ending points need to be stored, and the trajectory can be reconstructed by linear interpolation; 4) This trajectory is continuous, rather than consisting of intermittently sampled points like the original path, which greatly enriches our training set.

\begin{figure*}[t!]
  \centering
  \includegraphics[width=0.92\textwidth]{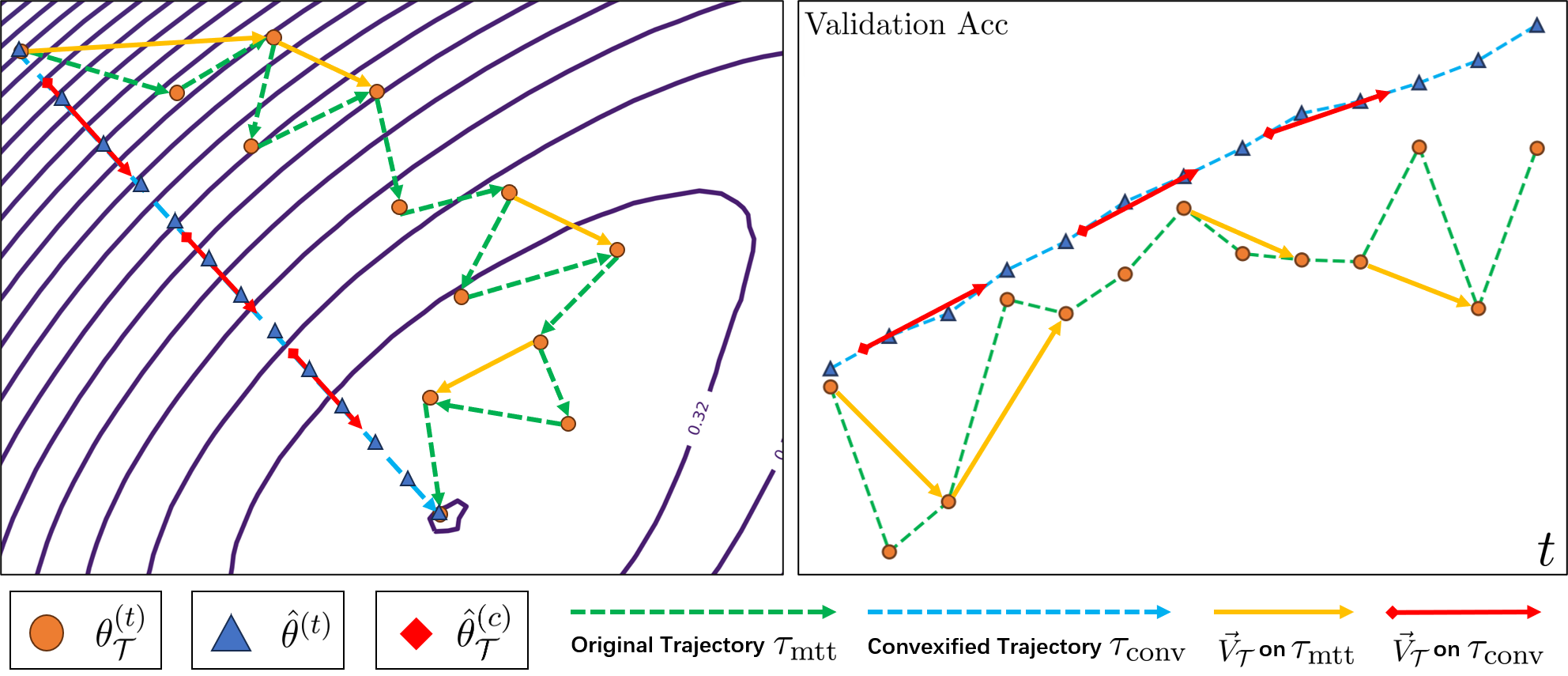}
  \caption{An illustration of the proposed MCT method. The left figure illustrates a schematic of the landscape in the model parameter space, while the right figure shows the validation accuracy of waypoint models extracted from expert trajectories of both the MTT method and our MCT method. In the left figure, the original trajectory $\tau_\text{mtt}$ exhibits constant oscillations, causing $\Vec{V}^{(t)}_{\mathcal{T}}$ to continuously change, resulting in fluctuating accuracy of the expert model in the right figure. In contrast, the trajectory $\tau_\text{conv}$ of our MCT method is very stable, thereby ensuring a consistent guidance direction, which leads to a steady improvement of the expert model as shown in the right figure.}
  \label{Fig:framework}
\end{figure*}

\section{Our proposed MCT Method}


\subsection{Matching Convexified Trajectory}
The expert trajectory $\tau_{\text{mtt}}$ is pre-generated with the parameter of all waypoint models stored in memory, \textit{i.e.}, $\tau_{\text{mtt}} = \{\theta^{(0)}_{\mathcal{T}}, \theta^{(1)}_{\mathcal{T}}, \dots, \theta^{(t)}_{\mathcal{T}}, \dots, \theta^{(K)}_{\mathcal{T}}\}$, where $\theta^{(t)}_{\mathcal{T}}$ is computed by multiple steps of mini-batch SGD \citep{cazenavette2022dataset}.

However, the trajectory generated by vanilla mini-batch SGD exhibits strong non-convexity, which makes synthetic data challenging to converge to an optimal solution. 
To this end, we proposed MCT, which creates a convexified trajectory $\tau_{\text{conv}}$ and is defined as: 
\begin{equation}
    \tau_{\text{conv}} = \{\hat{\theta}^{(t)}|0\leq t\leq K\},
\end{equation}
\begin{equation}
     \hat{\theta}^{(t)} = (1-\beta^{(t)}) {\theta}^{(0)}_{\mathcal{T}} + \beta^{(t)} {\theta}^{(K)}_{\mathcal{T}},
\end{equation}
where $\beta^{(t)} \in (0,1)$ is a weight value that determines the distribution of all waypoints. 
The starting point $\hat{\theta}^{(0)}$ and ending point $\hat{\theta}^{(K)}$ are same as ${\theta}^{(0)}_{\mathcal{T}}$ and ${\theta}^{(K)}_{\mathcal{T}}$ in $\tau_{\text{conv}}$. 
Particularly, the generated trajectory $\tau_{\text{conv}}$ directly points from ${\theta}^{(0)}_{\mathcal{T}}$ to ${\theta}^{(K)}_{\mathcal{T}}$, and  $\beta^{(t)}$ is determined by the ratio of the difference between ${\theta}^{(t-1)}_{\mathcal{T}}$ and ${\theta}^{(t)}_{\mathcal{T}}$ in $\tau_{\text{mtt}}$ to the total length of $\tau_{\text{mtt}}$ as:
\begin{align}
    \begin{split}
    \beta^{(0)} &= 0, \\
    \beta^{(t)} &= \frac{
        \sum_{l=0}^{t-1}{ 
            \mathrm{Norm}( \theta_{\mathcal{T}}^{(l+1)} - \theta_{\mathcal{T}}^{(l)} )
        }
    }{
        \sum_{l=0}^{K-1}{ 
            \mathrm{Norm}( \theta_{\mathcal{T}}^{(l+1)} - \theta_{\mathcal{T}}^{(l)} )
        }
    }, t = 1,2,\dots,K, 
    \end{split}
\end{align}
where $\mathrm{Norm}(\cdot)$ is $L_2$ norm. 
To mitigate discrepancies among different network layers, we calculate the $L_2$ normalization for each layer individually, \textit{i.e.}, $\beta^{(t)} = [\beta^{(t)}_1, \beta^{(t)}_2, \dots, \beta^{(t)}_n]^{\mathsf{T}}$, where each element in $\beta^{(t)}$ represents the weight value of a network layer. Note that our trajectory is generated based on $\tau_{\text{mtt}}$, and the calculation of $\beta^{(t)}$ does not require saving all the intermediate models ${\theta}^{(t)}_{\mathcal{T}}$. It only needs to save $\mathrm{Norm}( \theta_{\mathcal{T}}^{(l+1)} - \theta_{\mathcal{T}}^{(l)} )$ obtained in each step of the expert trajectory $\tau_{\text{mtt}}$, allowing $\beta^{(t)}$ to be calculated at the end of expert training. Given this expert trajectory $\tau_{\text{conv}}$, the distillation in Equ. \ref{equ:MTT_loss} can be conducted. During distillation, our MCT method always provides a convexified guidance $\Vec{{V}}^{(t)}_{\mathcal{T}}$ with the direction from ${\theta}^{(0)}_{\mathcal{T}}$ to ${\theta}^{(K)}_{\mathcal{T}}$, leading to the steady optimization of $\Vec{{V}}^{(t)}_{\mathcal{S}}$, and thus, the convergence of $\mathcal{S}$ will be rapid.


\subsection{Continuous Sampling}

Due to the continuity of our convexified trajectory, we can perform continuous sampling from the trajectory during distillation. This approach is completely different from the MTT method, enabling the selection of intermediate positions such as "the 1.5th point."
Specifically, the MTT method only performs discrete sampling on the expert trajectory (\textit{i.e.}, selecting ${\theta}^{(t)}_{\mathcal{T}}$ with an integer $t$ ). 
In contrast, for $\tau_{\text{conv}}$ with the starting point $\hat{\theta}^{(0)}$ and ending point $\hat{\theta}^{(K)}$, since all points are on a straight line, we can obtain any timestep $\hat{\theta}^{(c)}$ with a decimal $c \in [0,K]$ on this line by interpolation:
\begin{align}
    \begin{split}
        \hat{\theta}^{(c)} &= (1- \hat{\beta} ) \hat{\theta}^{(0)} + \hat{\beta} \hat{\theta}^{(K)}, \\
        \hat{\beta} &= (1-\eta) \beta^{(\lfloor c \rfloor)} + \eta \beta^{(\lceil c \rceil)}, \\
        \eta &= c - \lfloor c \rfloor, 
    \end{split}
    \label{equ:continous_sampling}
\end{align}
After $\hat{\theta}^{(c)}$ and $\hat{\theta}^{(c+M)}$ are obtained, the distillation process can be conducted. This continuous sampling strategy ensures sufficient learning and fitting of the entire expert trajectory $\tau_{\text{conv}}$, facilitating thorough learning of the synthetic dataset $\mathcal{S}$.   

\subsection{Memory-Efficient Storage}
\label{sec:memory_storage}
In conventional MTT, the learning of the expert trajectory requires storing the parameters of all timesteps in memory, which will incur significant storage overhead. Formally, let $W$ denote the size of the network parameter $\theta^{(t)}_\mathcal{T}$ and $C$ denote the size of other irrelevant parameters. Since there are $K$ timesteps on $\tau_{\text{mtt}}$, the entire required storage will be:
\begin{equation}
    \text{Storage}_\text{mtt} = K \times W + C = O(KW).   
\end{equation}

In contrast, our method only requires storing the starting point $\hat{\theta}^{(0)}$, the ending point $\hat{\theta}^{(K)}$, and point distribution $\{\beta^{(t)}|0 \leq t \leq K\}$ along the trajectory. Therefore, the entire storage cost becomes: 
\begin{equation}
    \text{Storage}_\text{conv} = 2 \times W + K \times (\beta^{(t)}) + C. 
\end{equation}
Since $\beta^{(t)}$ is a floating-point number, the storage cost will be $\text{Storage}_\text{conv} = O(W)$. In practice, $K$ is usually set to 50. 
Once the surrogate models in distillation become complex (\textit{e.g.} LLMs), $K$ and $W$ will increase simultaneously, highlighting the significant storage advantages of our MCT method.

\begin{figure}[t!]
  \centering
  \begin{subfigure}{0.30\linewidth}
    \includegraphics[width=\linewidth]{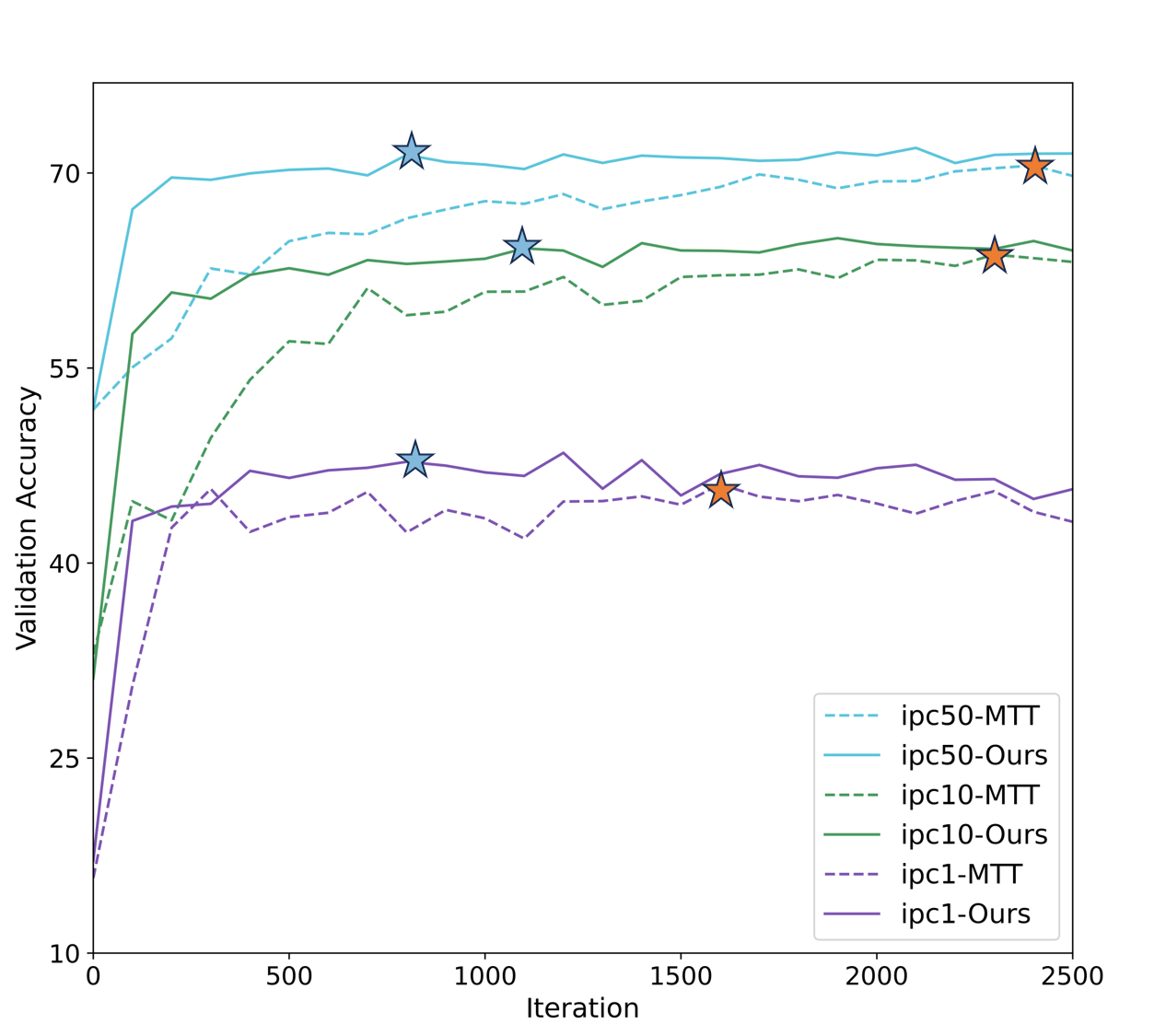}
    \caption{Convergence on CIFAR-10}
    \label{fig:CIFAR10_speed_compare}
  \end{subfigure}
  \begin{subfigure}{0.30\linewidth}
    \includegraphics[width=\linewidth]{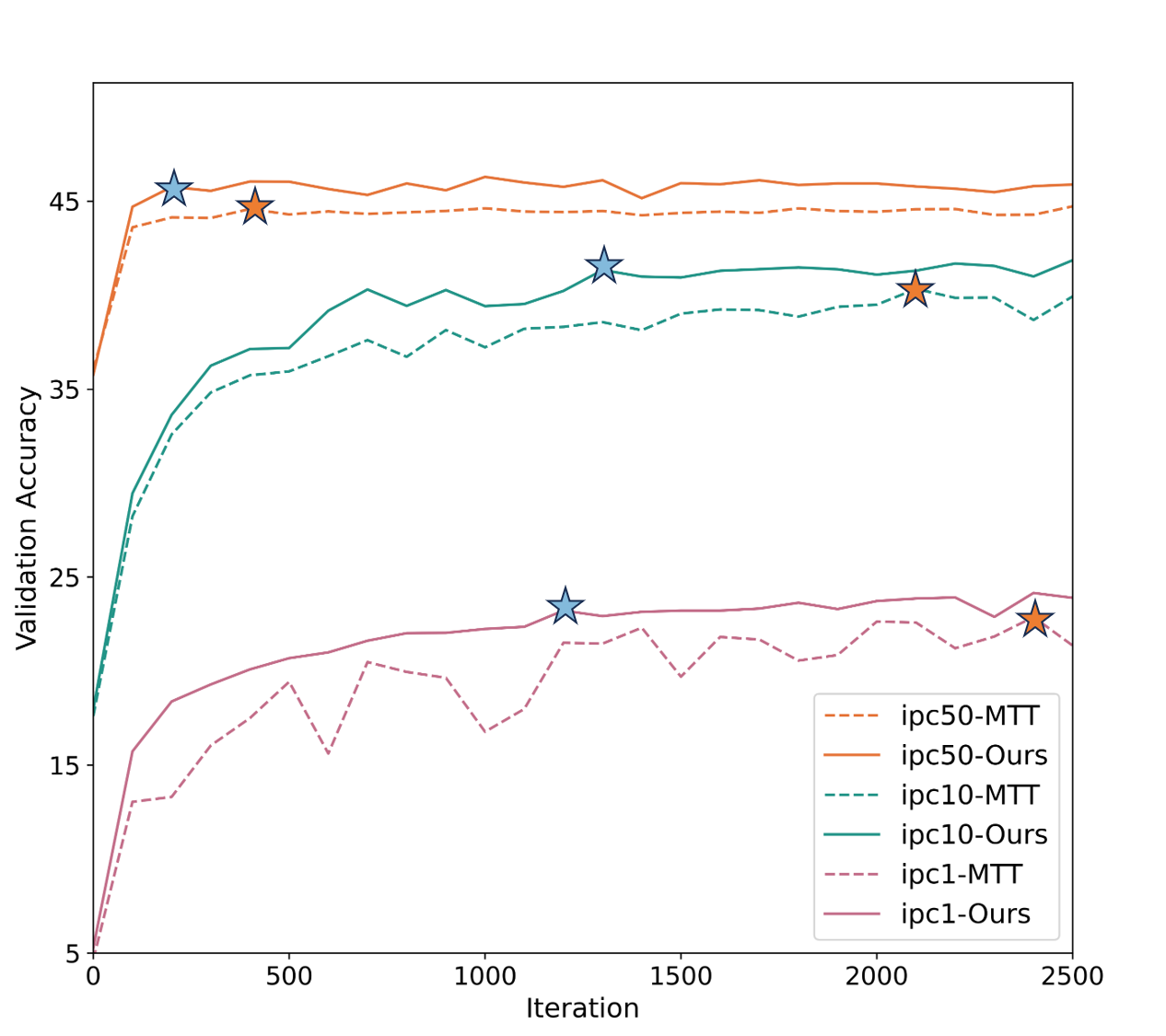}
    \caption{Convergence on CIFAR-100}
    \label{fig:CIFAR100_speed_compare}
  \end{subfigure}
  \begin{subfigure}{0.30\linewidth}
    \includegraphics[width=\linewidth]{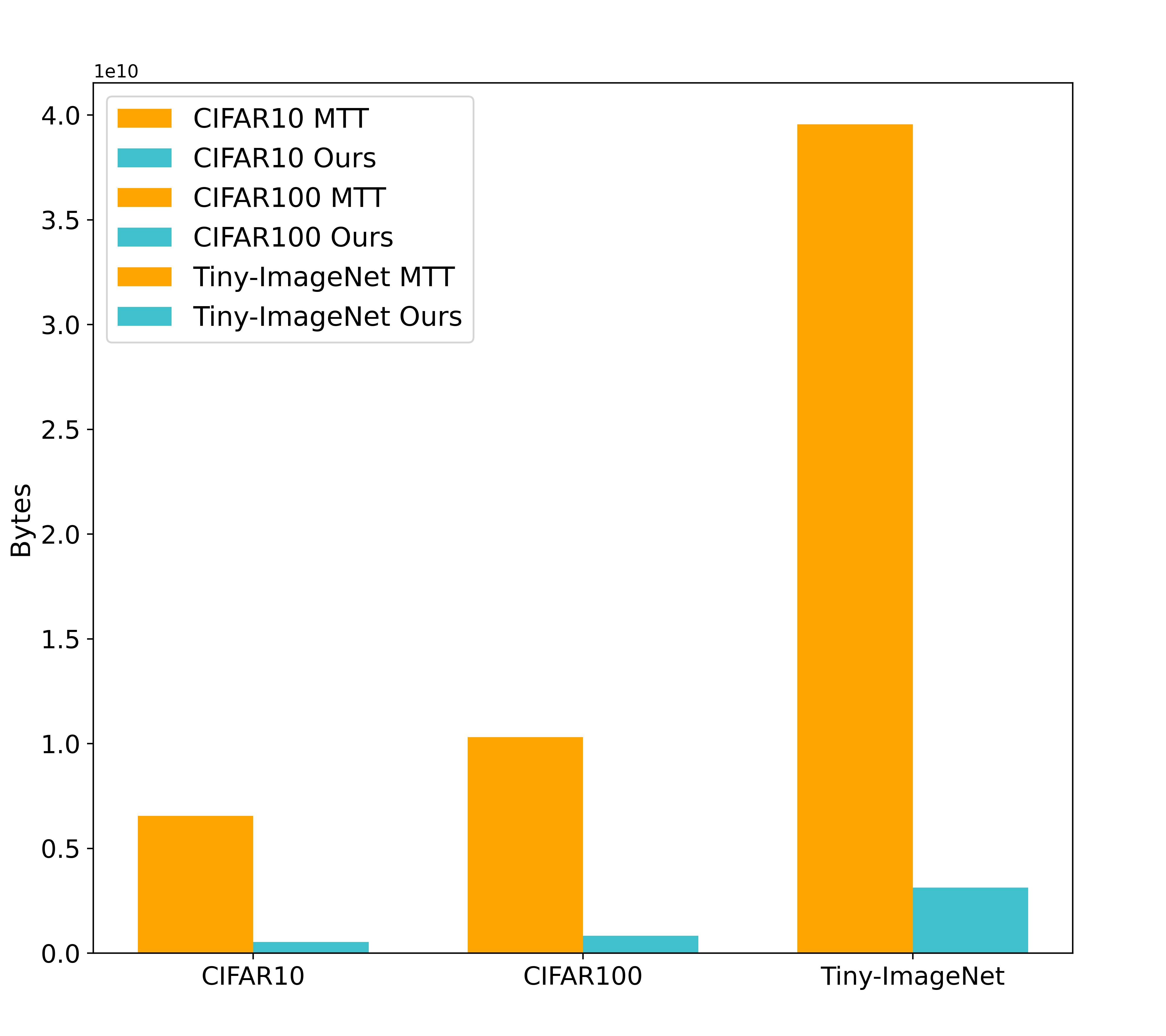}
    \caption{Storage Comparison}
    \label{fig:storage_compare}
  \end{subfigure}
  \caption{(a) and (b): Convergence comparisons of distillation process on CIFAR-10 and CIFAR-100, where the symbol ``star'' denotes the convergence point. (c): Storage comparisons on three datasets.}
  \label{conv}
\end{figure}
\section{Experiment}

\subsection{Experiment Setup}
\label{sec:experiment_setup}
\textbf{Experiment Settings:} 
We evaluated our method on three datasets: CIFAR-10 and CIFAR-100 \citep{krizhevsky2009learning}, and Tiny-ImageNet \citep{le2015tiny}. 
We first generated the convexified trajectories with our MCT method. 
Similar to MTT, we applied Kornia \citep{riba2020kornia} Zero component analysis (ZCA) whitening on CIFAR-10, CIFAR-100, and Tiny-ImageNet datasets, and utilized Differentiable Siamese Augmentation (DSA) \citep{zhao2021dataset} technique during training and evaluation.

\textbf{Evaluation and Baselines:} 
Our MCT method is compared with several baselines from different branches, including Dataset Condensation (DC) \citep{zhao2020dataset}, Distribution Matching (DM) \citep{zhao2023dataset},  DSA \citep{zhao2021dataset}, Condense Aligning FEatures (CAFE) \cite{wang2022cafe}, dataset distillation using Parameter Pruning (PP) \citep{li2023dataset}, and MTT. Following the conventional settings, we conducted dataset distillation using 1/10/50 images per class (ipc) for evaluations, respectively. The images with the resolution of 32 × 32 and 64 × 64 were synthesized on the CIFAR and Tiny-ImageNet datasets, respectively.
Subsequently, five randomly initialized networks were trained in 1000 iterations with the cross-entropy loss on the distilled dataset. These trained networks were then evaluated on the real validation set, and their average accuracy (Acc) was reported as the evaluation metric. To maintain consistency with MTT and DC, we use ConvNet \citep{gidaris2018dynamic} as the surrogate model. This model comprises 128 filters with a 3 × 3 kernel size. Following the filters, instance normalization \citep{ulyanov2016instance} and ReLU activation are applied. Additionally, an average pooling layer with a kernel size of 2 × 2 and a stride of 2 is incorporated into the network.

\textbf{Implementation Details:}
We adopt the same settings of MTT in most cases. Specifically, 100 expert trajectories are generated, each spanning 50 epochs of training (\textit{i.e.}, 51 timesteps).  In practice, we often insert two waypoint models in the expert trajectory of MTT to derive our convexified trajectory:  the models of 6-th and 25-th epochs for CIFAR-10 and the models of 15-th and 30-th epochs for CIFAR-100 and Tiny-ImageNet. During the distillation process, 5,000 distillation iterations are conducted. For each iteration, $\hat{\theta}^{(c)}$ is generated from Equ.~\ref{equ:continous_sampling}, where the decimal $c$ is randomly sampled within [0, MaxStartEpoch]. We adopt the SGD optimizer, and a learnable learning rate is employed to distill the synthetic data. All experiments are run on four RTX3090 GPUs.

\begin{table}[t!]
\centering
\caption{Performance of Various Algorithms on Different Datasets}
\label{tab:performance}
\resizebox{0.97\linewidth}{!}{
\begin{tabular}{l|ccc|ccc|ccc}
    \hline
    Dataset & \multicolumn{3}{c|}{CIFAR-10} & \multicolumn{3}{c|}{CIFAR-100} & \multicolumn{3}{c}{Tiny ImageNet}\\
 
    ipc & 1 & 10 & 50 & 1 & 10 & 50 & 1 & 10 & 50 \\
    \hline
    Random & 15.4±0.3 & 31.0±0.5 & 50.6±0.3 & 4.2±0.3 & 14.6±0.5 & 33.4±0.4 & 1.4±0.1 & 5.0±0.2 & 15.0±0.4 \\
    DC \citep{zhao2020dataset} & 28.3±0.5 & 44.9±0.5 & 53.9±0.5 & 12.8±0.3 & 25.2±0.3 & - & - & - & - \\
    DSA \citep{zhao2021dataset} & 28.8±0.7 & 52.1±0.5 & 60.6±0.5 & 13.9±0.3 & 32.3±0.3 & 42.8±0.4 & - & - & - \\
    CAFE \citep{wang2022cafe} & 30.3±1.1 & 46.3±0.6 & 55.5±0.6 & 12.9±0.3 & 27.8±0.3 & 37.9±0.3 & - & - & - \\
    DM \citep{zhao2023dataset} & 26.0±0.8 & 48.9±0.6 & 63.0±0.4 & 11.4±0.3 & 29.7±0.3 & 43.6±0.4 & 3.9±0.2 & 12.9±0.4 & 24.1±0.3 \\
    PP \citep{li2023dataset} & \textit{46.4±0.6} & \textit{65.5±0.3} & \textit{71.9±0.2} & \textbf{24.6±0.1} & \textbf{43.1±0.3} & \textbf{48.4±0.3} & - & - & - \\
    MTT \citep{cazenavette2022dataset} & 46.3±0.8 & 65.3±0.7 & 71.6±0.2 & 24.3±0.3 & 40.1±0.4 & \textit{47.7±0.2} & \textit{8.8±0.3} & \textbf{23.2±0.2} & \textbf{28.0±0.3} \\
    Ours & \textbf{48.5±0.2} & \textbf{66.0±0.3} & \textbf{72.3±0.3} & \textit{24.5±0.5} & \textit{42.5±0.5} & 46.8±0.2 & \textbf{9.6±0.5} & \textit{22.6±0.8} & \textit{27.6±0.4} \\
     \hline
    Full dataset & \multicolumn{3}{c|}{84.8±0.1} & \multicolumn{3}{c|}{56.2±0.3} & \multicolumn{3}{c}{37.6±0.4} \\
    \hline
\end{tabular}
}
\label{tab:main_result}
\end{table}

\subsection{Experiment Result}

\textbf{Validation Accuracy Comparison.} Table \ref{tab:main_result} presents a comparison of validation accuracy between our method and various baselines across three datasets. Although performance is not the main focus of our MCT method, it is evident that our method achieves the best performance on the three metrics of the CIFAR-10 dataset as well as the ipc=1 metric of the Tiny ImageNet dataset. Notably, compared to the crucial MTT method, our MCT method demonstrates performance improvements in most metrics, indicating that our convexified trajectory and continuous sampling strategy can indeed provide enhanced guidance to the optimization of synthetic datasets.

\textbf{Convergence of Distillation Process.}
Figure \ref{fig:CIFAR10_speed_compare} and \ref{fig:CIFAR100_speed_compare} illustrate the distillation processes utilizing the MCT and MTT methods for the CIFAR-10 and CIFAR-100 datasets. After every 100 distillation iterations, five networks with random initialization are trained on the current distillation dataset and their average accuracy on the validation set are recorded. The figures present the validation accuracy trends of both methods over the initial 2,500 iterations. As depicted, under all ipc settings, our MCT method achieves a substantial performance much sooner (200-1200 iterations ahead), indicating a faster convergence speed; after nearing convergence, the performance of the MCT method remains consistently stable as iterations proceed, whereas the MTT method still experiences significant performance fluctuations. These two phenomena suggest that our method effectively enhances training stability and accelerates the convergence process.



\textbf{Comparison of Storage Requirement.}
Figure \ref{fig:storage_compare} compares the required storage of the expert trajectory between MTT and our MCT method. As demonstrated in Sec. \ref{sec:memory_storage}, it is clear that our convex trajectories require significantly less memory (approximately 8\%) compared to the expert trajectories needed by the MTT method. It is foreseeable that as model sizes and expert trajectories continue to grow, the space savings offered by our method will become even more substantial.

\begin{figure}
  \centering
  \begin{subfigure}{0.45\linewidth}
    \includegraphics[width=\linewidth]{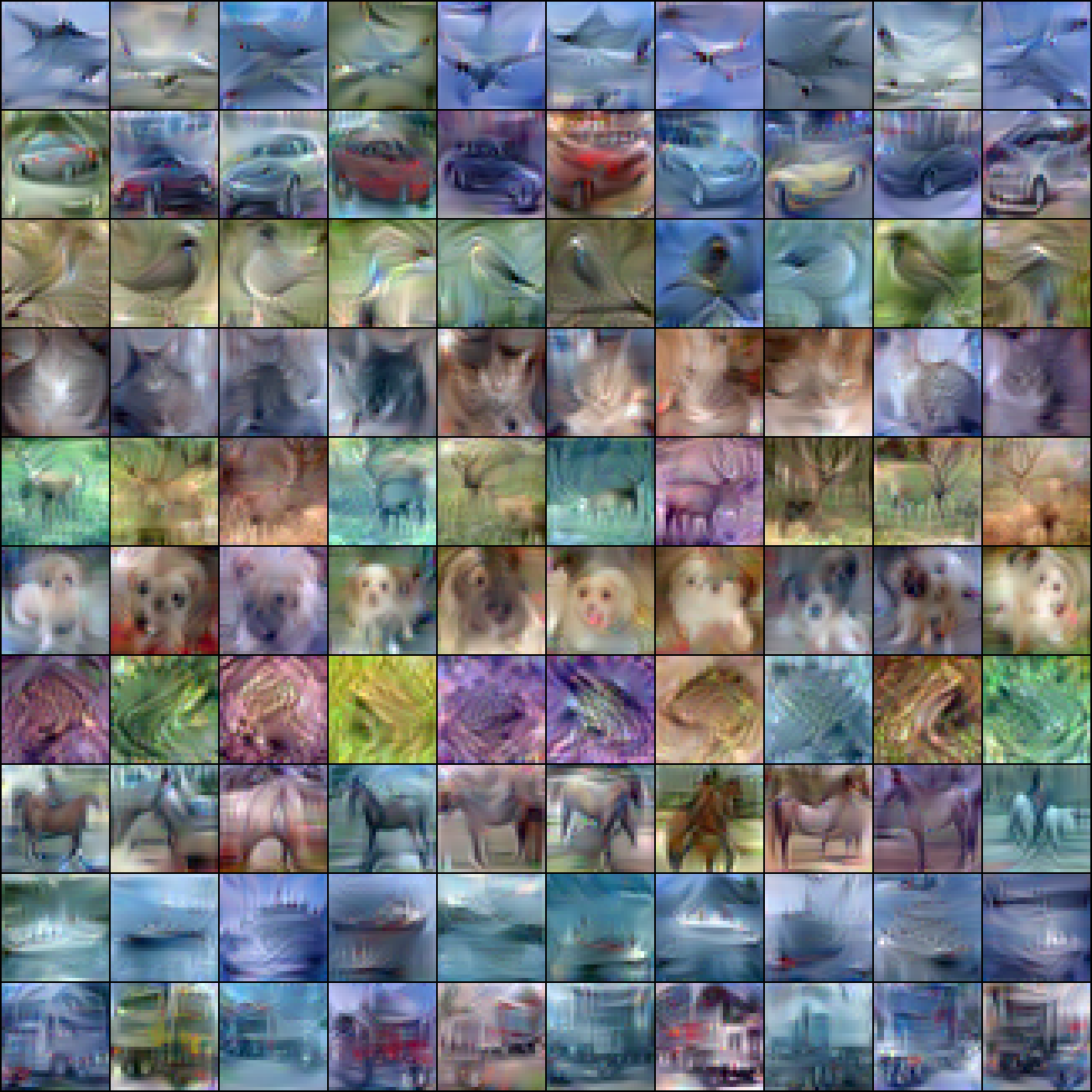}
    \caption{CIFAR-10, ipc=10}
    \label{fig:visual_cf10}
  \end{subfigure}
  \begin{subfigure}{0.45\linewidth}
    \includegraphics[width=\linewidth]{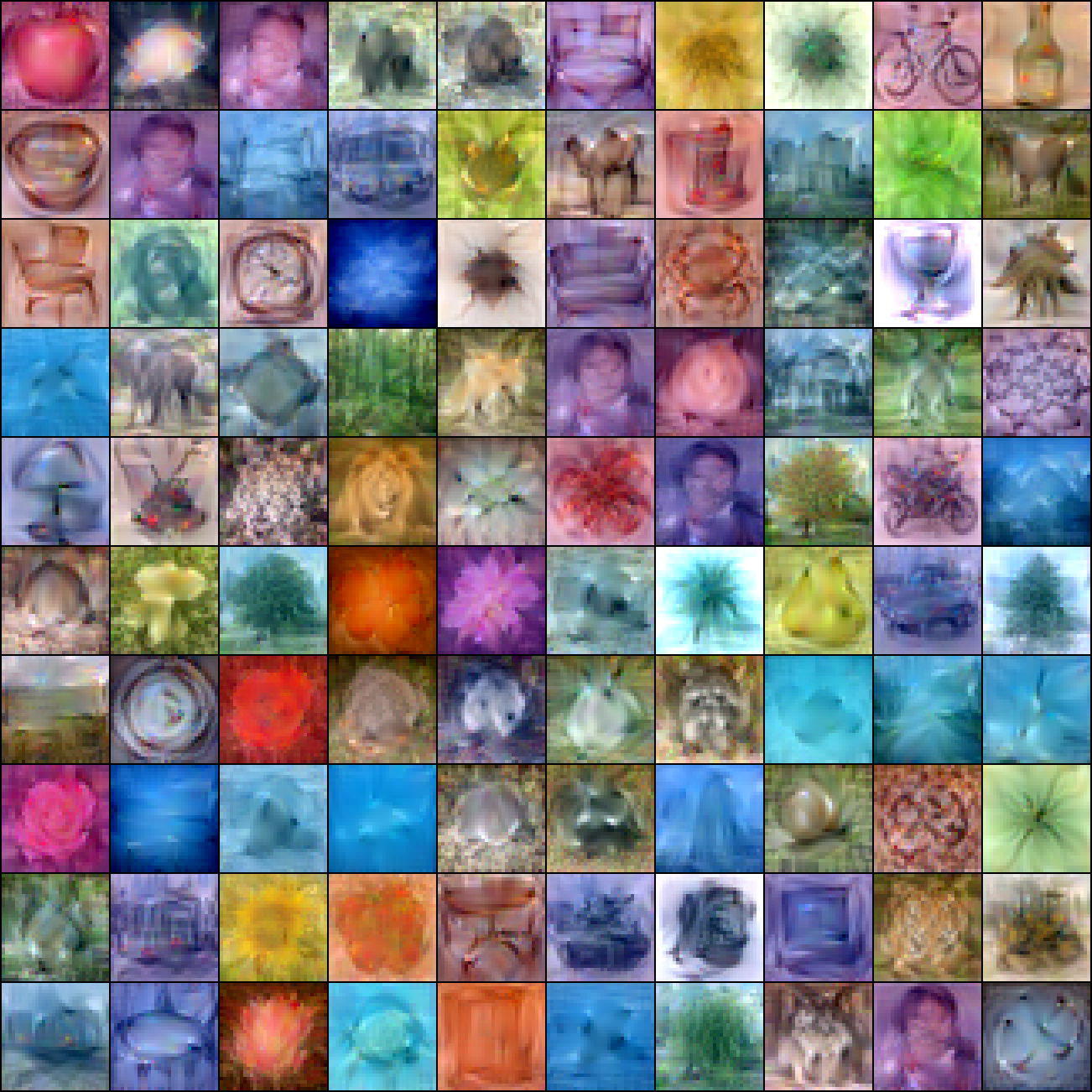}
    \caption{CIFAR-100, ipc=1}
    \label{fig:visual_cf100}
  \end{subfigure}
  \caption{Visualization of synthetic dataset. }
  \vspace{-5.5pt}
  \label{conv}
\end{figure}

\textbf{Visualization of Distilled Data.}
The visualization results of the synthetic data on CIFAR-10 with ipc=10 and CIFAR-100 with ipc=1 are presented in Figure~\ref{fig:visual_cf10} and \ref{fig:visual_cf100}. As we can see, the synthetic set learned from our expert trajectories exhibits notable degrees of recognizability and authenticity, while it also tends to integrate various characteristic features of images within the same category.


\begin{figure}
  \centering
  \begin{subfigure}{0.30\linewidth}
    \includegraphics[width=\linewidth]{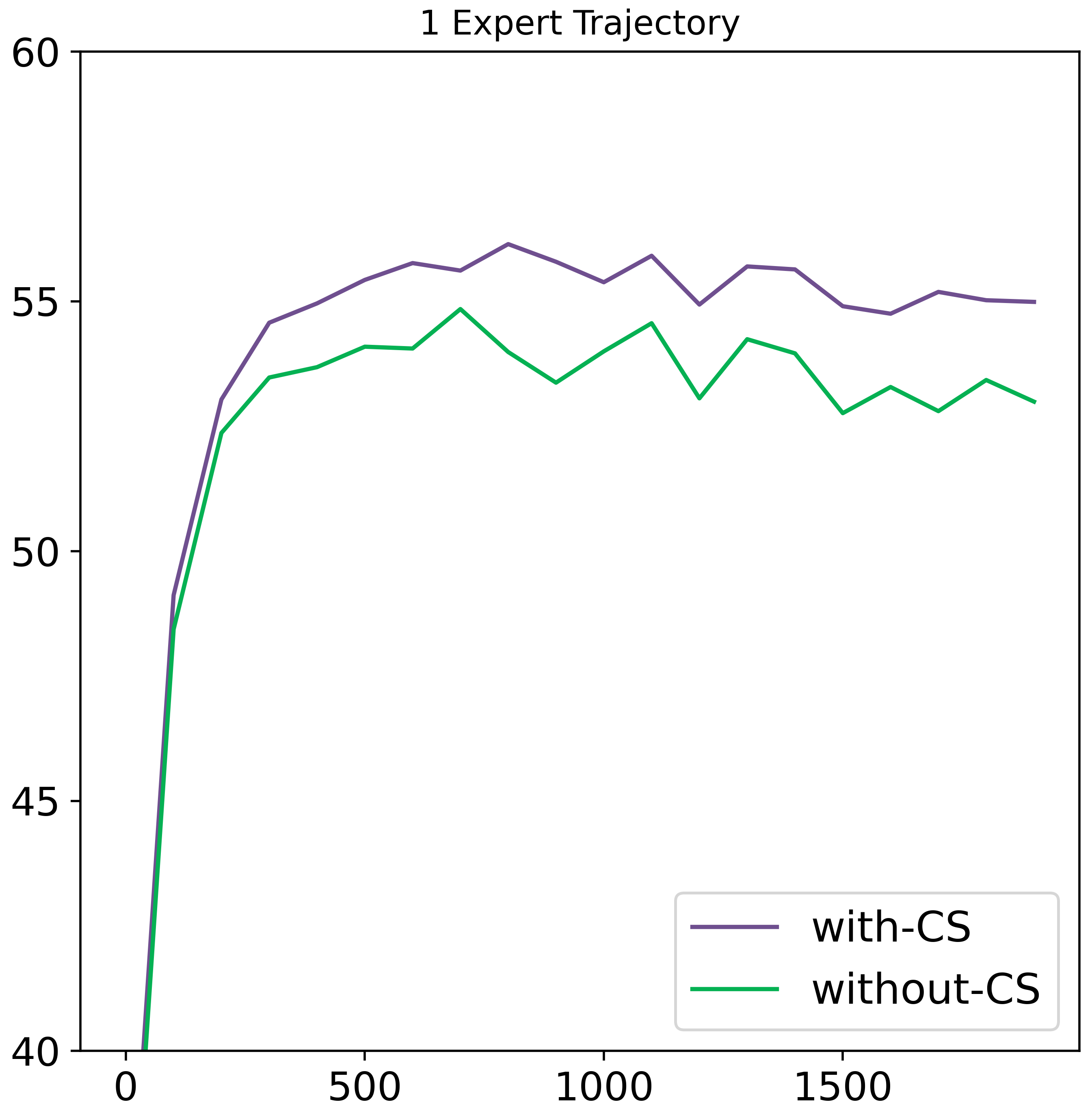}
    \caption{1 Expert Trajectory}
    \label{fig:expert_num_1}
  \end{subfigure}
  \begin{subfigure}{0.30\linewidth}
    \includegraphics[width=\linewidth]{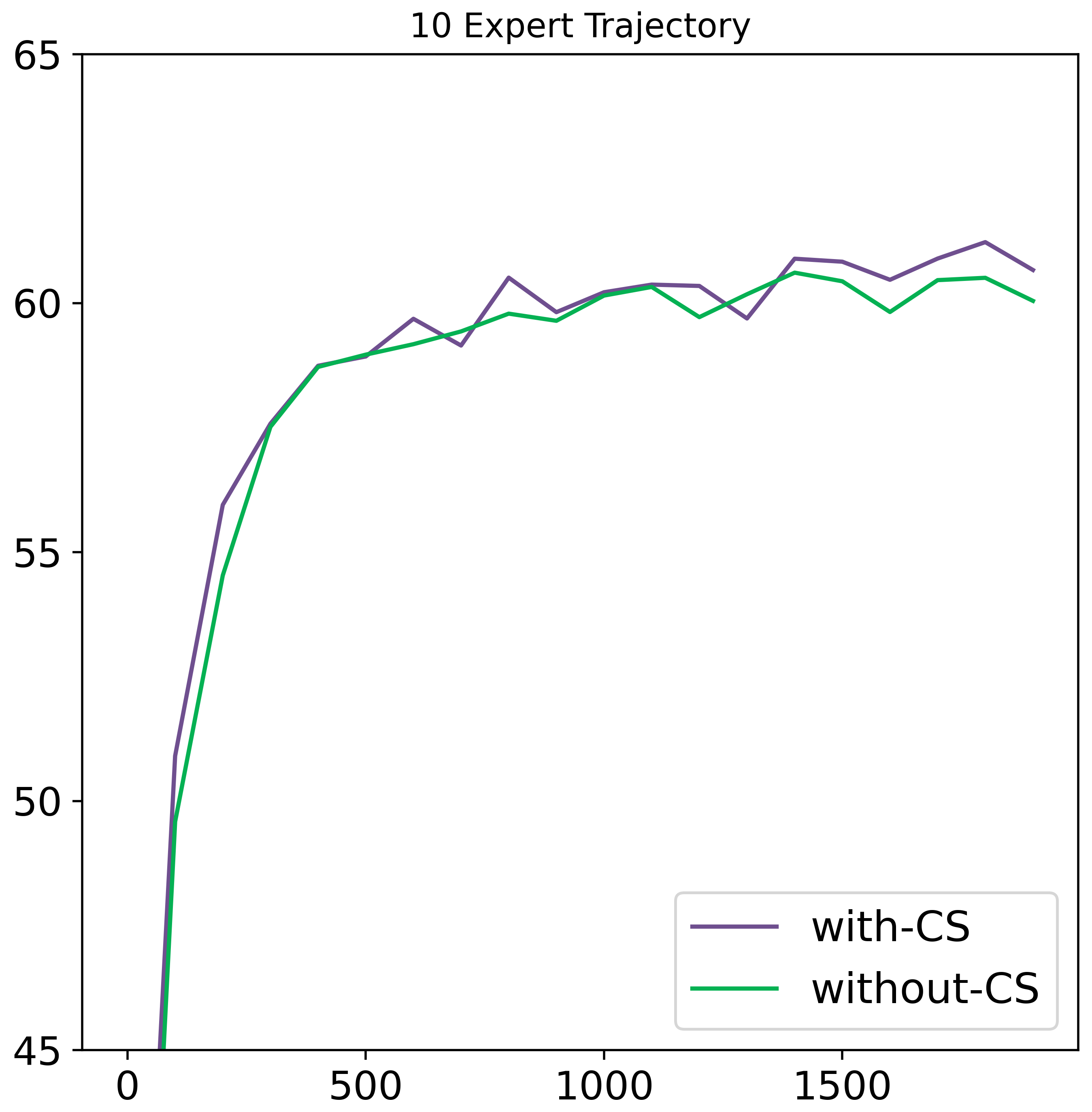}
    \caption{10 Expert Trajectories}
    \label{fig:expert_num_10}
  \end{subfigure}
  \begin{subfigure}{0.30\linewidth}
    \includegraphics[width=\linewidth]{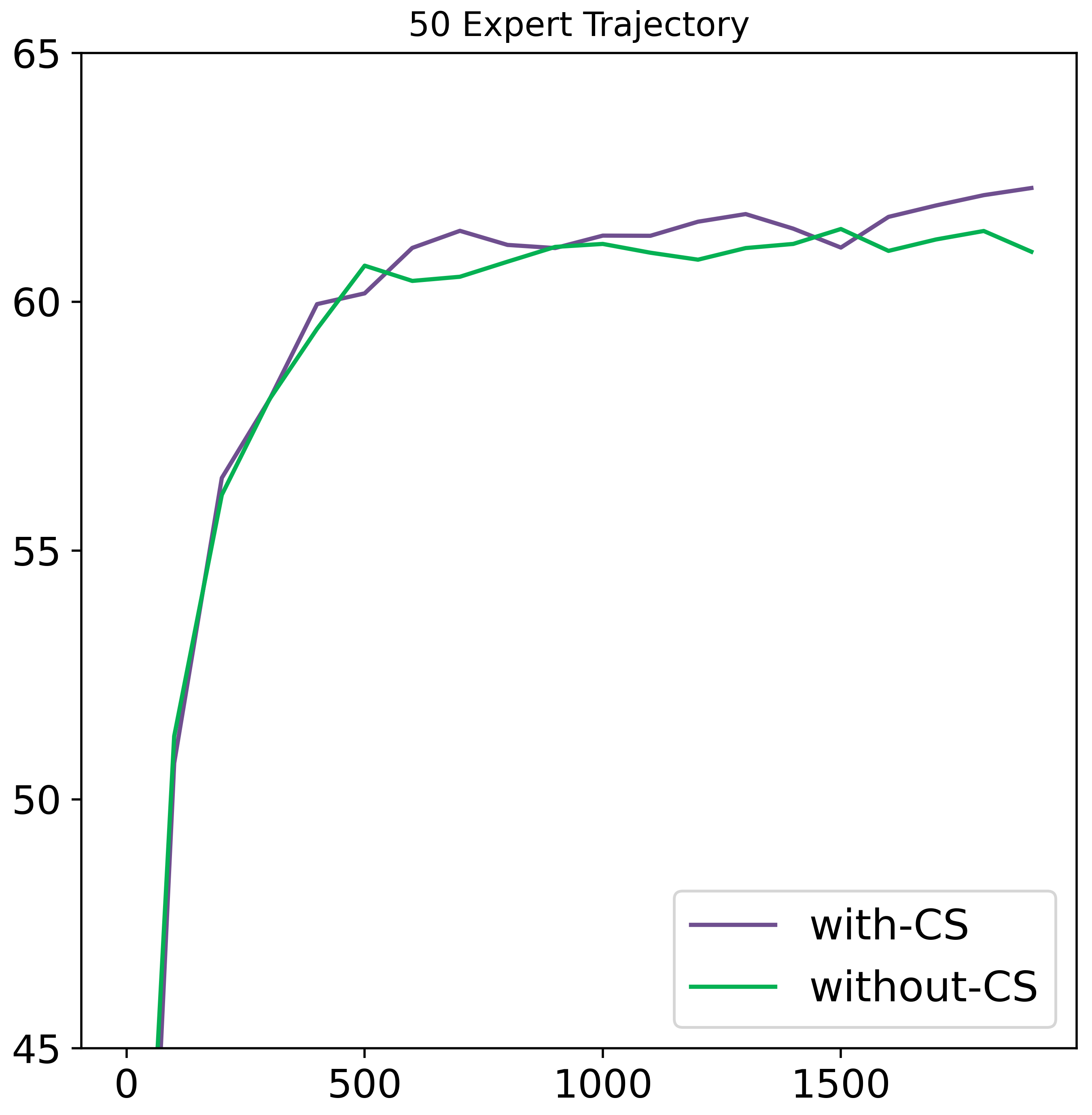}
    \caption{50 Expert Trajectories}
    \label{fig:expert_num_50}
  \end{subfigure}

  \caption{{Effects of Continuous Sampling over iterations with different expert trajectory numbers.} }
  \vspace{-9pt}
  \label{fig:expert_num}
\end{figure}

\subsection{Ablation Studies}


\begin{table}[t!]
    \centering
    \caption{{Effects of Continuous Sampling with different numbers of expert trajectories on CIFAR-10.} }
    \begin{tabular}{c|ccccc}
        \hline
        \textbf{Number of expert trajectories} & \textbf{1} & \textbf{5} & \textbf{10} & \textbf{20}& \textbf{50} \\ \hline
        \textbf{w/o. Continuous Sampling}        & 54.8±0.2       & 60.6±0.2      & 61.5±0.3   &  62.3±0.3 &   62.1±0.4\\ 
        \textbf{w. Continuous Sampling}         & 56.2±0.3       & 61.3±0.5       & 61.8±0.6  &  62.8±0.3  &  62.8±0.2\\ \hline
    \end{tabular}
    \label{tab:continuous_sampling}
\end{table}

\begin{table}[t!]
    \centering
    \caption{{Effects of different $M$ with different ipc on CIFAR-10.}}
    \begin{tabular}{c|ccccc}
        \hline
        \textbf{$M$}  & \textbf{3} & \textbf{4} & \textbf{5} & \textbf{6} & \textbf{7} \\ \hline
        \textbf{ipc=1}     & 46.7      & 47.1      & \textbf{48.5}      & 48.0 & 45.6     \\ \hline
        \textbf{ipc=10}    & 62.3      & 62.6      & 65.0      & \textbf{66.0 }& 65.2      \\ \hline
        \textbf{ipc=50}    & 70.0      & 71.4      & 71.8      & \textbf{72.3} & 71.8     \\ \hline
    \end{tabular}
    \label{tab:effect_of_m}
\end{table}

\textbf{Effects of Continuous Sampling.}
To verify the effect of the continuous sampling, we set ipc=10 and randomized the starting epoch parameter within the range [0,5] on the CIFAR-10 dataset. The validation accuracy over iterations and the optimal accuracy throughout the entire distillation process are reported in Figure \ref{fig:expert_num_50} and Table. \ref{tab:continuous_sampling}, respectively. Overall, the integration of continuous sampling can improve the validation performance under all conditions. Moreover, the fewer the number of expert trajectories, the more pronounced the performance improvement brought about by the continuous sampling strategy. Those results prove that our continuous sampling can effectively expand the sampling space, ultimately leading to the enhancement of the final distillation outcomes.

\textbf{Effects of expert updating step $M$.}
Table \ref{tab:effect_of_m} shows the effects of the updating step $M$ of the expert trajectory $\tau_{\text{conv}}$ on the CIFAR-10 dataset. $N$ is set to 50 for all results.  As we can see, when ipc=1, the optimal performance can be obtained at $M$=5, while when ipc=10 and ipc=50, the optimal performance can be obtained at $M$=6. Overall, our MCT method is robust to the selection of $M$ and will not experience significant performance degradation with changes in $M$.

\section{Conclusion}
To address three major limitations of traditional MTT, this paper draws inspiration from NTK methods and proposes a novel perspective to understand the essence of dataset distillation and MTT. A simple yet novel Matching Convexified Trajectory method is introduced to create a simplified, convexified expert trajectory that enhances the optimization process, leading to more stable and rapid convergence and reduced memory consumption. The convexified trajectory allows for continuous sampling during distillation, enriching the learning process and ensuring thorough fitting of the expert trajectory.  Our experiments on CIFAR-10, CIFAR-100, and Tiny-ImageNet datasets demonstrate MCT's superiority over MTT and other baselines. MCT's ablation studies confirm the benefits of continuous sampling and the impact of the convexified trajectory on distillation performance. The results indicate that MCT is a promising solution for training complex models with reduced data needs, offering an efficient, stable, and memory-friendly approach to dataset distillation.


\section{Limitations}
\label{sec:limitation}
Our MCT method has three primary limitations: 1. Although MCT can effectively enhance training stability and convergence speed, the improvement in validation accuracy is not very significant due to the starting and ending points being the same as those in MTT, and the enhancement is mainly attributed to the more thorough trajectory learning enabled by continuous sampling; future work could identify better endpoints to further improve performance. 2. While our trajectory provides a direction with more stable and rapidly descending, the calculation the magnitude $\beta$ is relatively simple (derived by the proportion of MTT step size to trajectory length), and there may exist more optimal step sizes that allow for more rapid and robust trajectory learning. 
3. The linear approximation of NTK is proposed based on infinitely wide networks and requires some rather strict initialization methods. However, we did not conduct our work under this condition; further theoretical deduction is required to address this issue.

\bibliographystyle{plainnat}
\bibliography{dd}

\end{document}